\documentclass[letterpaper, 10 pt, conference]{ieeeconf}  

\IEEEoverridecommandlockouts                              
\usepackage{float}
\usepackage{graphicx}
\usepackage{lipsum}
\usepackage{stfloats}
\usepackage{multicol}
\usepackage{multirow}
\usepackage{amsmath,bm}
\usepackage{etoolbox}

\usepackage{array}

\usepackage{tabulary}
\usepackage{xcolor}

\usepackage{hyperref}
\hypersetup{colorlinks,linkcolor=black}
\hypersetup{colorlinks,urlcolor=black}
\hypersetup{colorlinks,citecolor=black}

\usepackage[caption=false,font=footnotesize]{subfig} 

\usepackage{makecell}

\usepackage{amsfonts}

\usepackage{pifont}

\title{\LARGE \bf
Panoramic Panoptic Segmentation: Towards Complete Surrounding Understanding via Unsupervised Contrastive Learning
}

\author{Alexander Jaus$^{1}$, Kailun Yang$^{1}$ and Rainer Stiefelhagen$^{1}$
\thanks{This work was supported in part by the Federal Ministry of Labor and Social Affairs (BMAS) through the AccessibleMaps project under Grant 01KM151112, in part by the University of Excellence through the ``KIT Future Fields'' project, and in part by the Hangzhou SurImage Technology Company Ltd. (\textit{Corresponding author: Kailun Yang}).}
\thanks{$^{1}$The authors are with Institute for Anthropomatics and Robotics, Karlsruhe Institute of Technology, 76131 Karlsruhe, Germany (e-mail: alexander.jaus@student.kit.edu, \{kailun.yang, rainer.stiefelhagen\}@kit.edu).}
}

\begin{document}

\maketitle
\thispagestyle{empty}
\pagestyle{empty}

\begin{abstract}

In this work, we introduce panoramic panoptic segmentation as the most holistic scene understanding both in terms of field of view and image level understanding for standard camera based input. A complete surrounding understanding provides a maximum of information to the agent, which is essential for any intelligent vehicle in order to make informed decisions in a safety-critical dynamic environment such as real-world traffic. In order to overcome the lack of annotated panoramic images, we propose a framework which allows model training on standard pinhole images and transfers the learned features to a different domain. Using our proposed method, we manage to achieve significant improvements of over 5\% measured in PQ over non-adapted models on our Wild Panoramic Panoptic Segmentation (WildPPS) dataset. We show that our proposed Panoramic Robust Feature (PRF) framework is not only suitable to improve performance on panoramic images but can be beneficial whenever model training and deployment are executed on data taken from different distributions. As an additional contribution, we publish WildPPS: The first panoramic panoptic image dataset to foster progress in surrounding perception.

\end{abstract}

\section{Introduction}
Computer vision has seen staggering improvements over the last years. Starting from a very coarse scene level understanding such as in image classification, state-of-the-art approaches provide for a very detailed understanding on a pixel level~\cite{long2015fully}. With the introduction of the new task of panoptic segmentation~\cite{kirillov2019panoptic} uniting semantic and instance segmentation, the so far most holistic scene level understanding can be achieved. Another approach to improve holistic scene understanding is to go beyond classical pinhole camera images. Panoramic images overcome the problem of a limited Field of View (FoV) and provide a more complete image of the real world~\cite{yang2019pass}. This additional information about the surrounding is critical for many real-world applications such as autonomous driving and various other robotic navigation tasks. In particular, Intelligent Vehicles (IV) require an in-depth understanding of the surrounding world~\cite{narioka2018understanding}. 

\begin{figure}[t]
    \centering
    \includegraphics[width=0.485\textwidth]{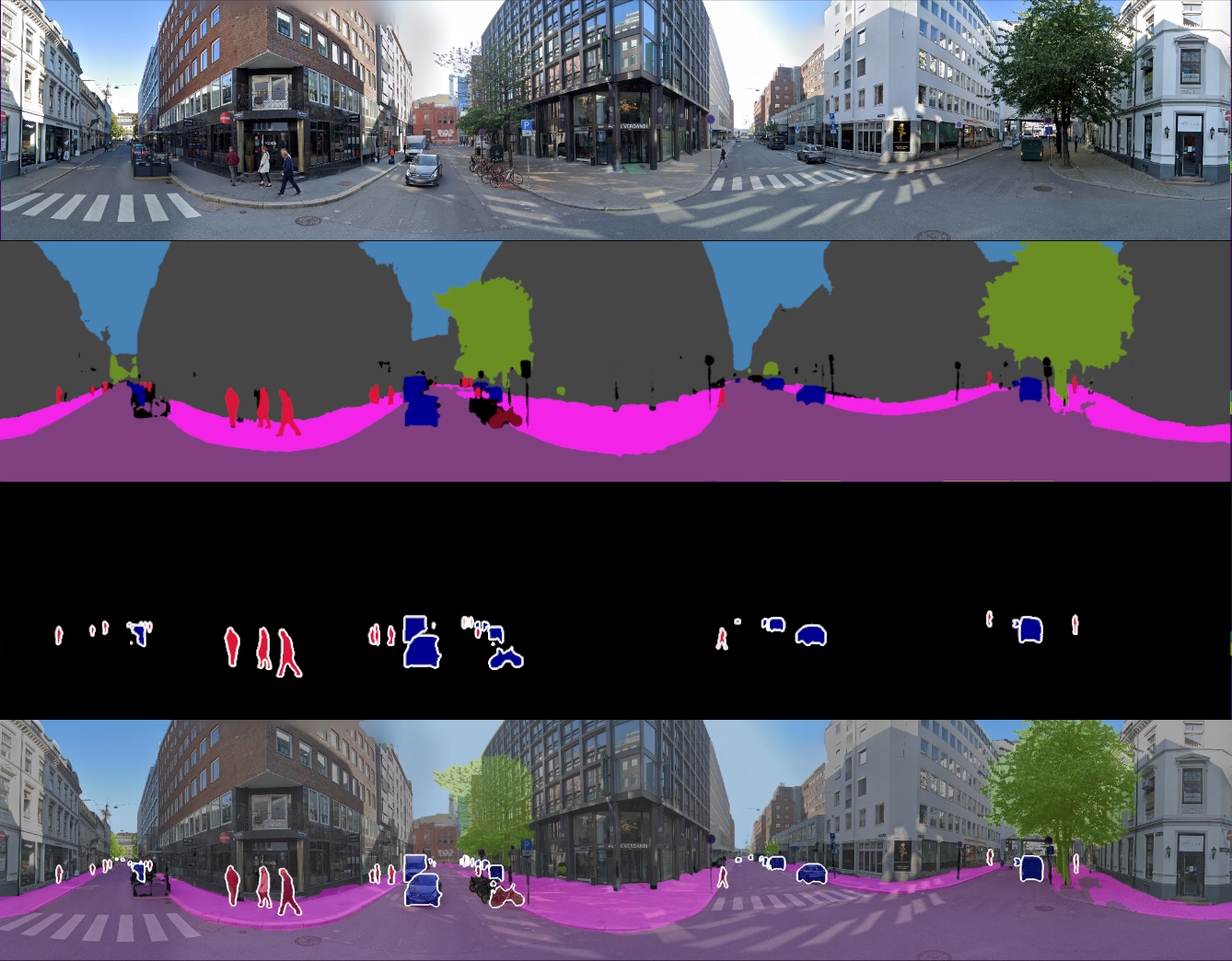}
    \vskip-1ex
    \caption{For a given panoramic image (first row), we define three levels of comprehension. The panoramic semantic understanding (second row) assigns a label to each pixel but does not distinguish between different instances of countable objects such as cars. The panoramic instance understanding (third row) detects each countable object and assigns a unique id and a pixel mask. Our proposed panoramic panoptic understanding (fourth row) is the so far most complete surrounding understanding by assigning a label to each pixel but also distinguishing between different instances.}
    \label{fig: panoptic panorama images}
    \vskip-3ex
\end{figure}

We argue that both, panoptic image level understanding and a 360$^\circ$ view are essential information for intelligent vehicles, especially those operating within a safety-critical and dynamic environment such as real-world traffic. Semantic segmentation results which provide information about uncountable stuff classes such as \textit{road} and \textit{sidewalk} are the basic building blocks of any decision making for intelligent vehicles. The additional information from distinguishing between different instances of countable classes such as \textit{car} and \textit{human} allow a more informed and thus safer interaction with the real world. \break 
A field of view beyond classical pinhole images is crucial for any dynamic environment, otherwise objects of interest can move out of sight~\cite{yang2019pass}.
Using multiple cameras or cameras along with other sensors such as LiDAR could be one solution to overcome this problem~\cite{narioka2018understanding}\cite{berrio2020camera}. However, it requires multiple different systems to work in parallel and combine those information to a holistic picture~\cite{berrio2020camera}, which incurs higher latency and computation complexity that are clearly not advantageous for intelligent vehicle applications. To address this issue, we pioneer to achieve panoramic panoptic segmentation, a purely vision-based task which is an arguably much simpler and more cost-efficient approach, since we parse all available information at once and output the so far most complete understanding of the surrounding based on 360$^\circ$ camera input (see Fig.~\ref{fig: panoptic panorama images}).

Lacking sufficient annotated data to directly train on panoramic images, we propose a novel approach to learn robust feature representations, allowing the model to better generalize on data from a different distribution. We experiment with various setups and find that augmenting the supervised training pipeline of our segmentation model by a short unsupervised pretraining step of the backbone using a combination of a contrastive loss and the pixel propagation loss introduced in~\cite{xie2020propagate} improves the performance up to $5.4\%$ in panoptic quality measure over the non-adapted baseline model on WildPPS. We furthermore observe that the gains of our approach are larger if the train and test datasets are more dissimilar, which is in line with our expectations that model predictions benefit more from robust features if train and test data are less alike.

The contributions of our paper are three-fold: 
We are the first to achieve panoptic segmentation on panoramic images, providing the so far most complete holistic scene understanding in both field of view and image level understanding.
Second, we introduce the Panoramic Robust Feature framework (PRF) as an approach which creates robust feature representations for the panoptic image segmentation downstream task by using recent unsupervised pretraining approaches~\cite{xie2020propagate}. We show that our learned feature representations are more robust compared to standard training procedures and thus show significant improvements for classes which substantially differ in looks between training and previously unseen testing set domains.

Third, in order to overcome the lack of annotated data to measure the panoptic performance of models on panoramic data, we provide the community with the first panoptic panoramic evaluation set consisting of fine labeled panoramic street images from 40 cities located on all continents. To foster future research in panoramic scene perception, all data will be made publicly available at.\footnote{PPS:~\url{https://github.com/alexanderjaus/PPS}}

\begin{figure}
    \centering
    \includegraphics[trim=0 60 0 60, clip, width=0.4\textwidth]{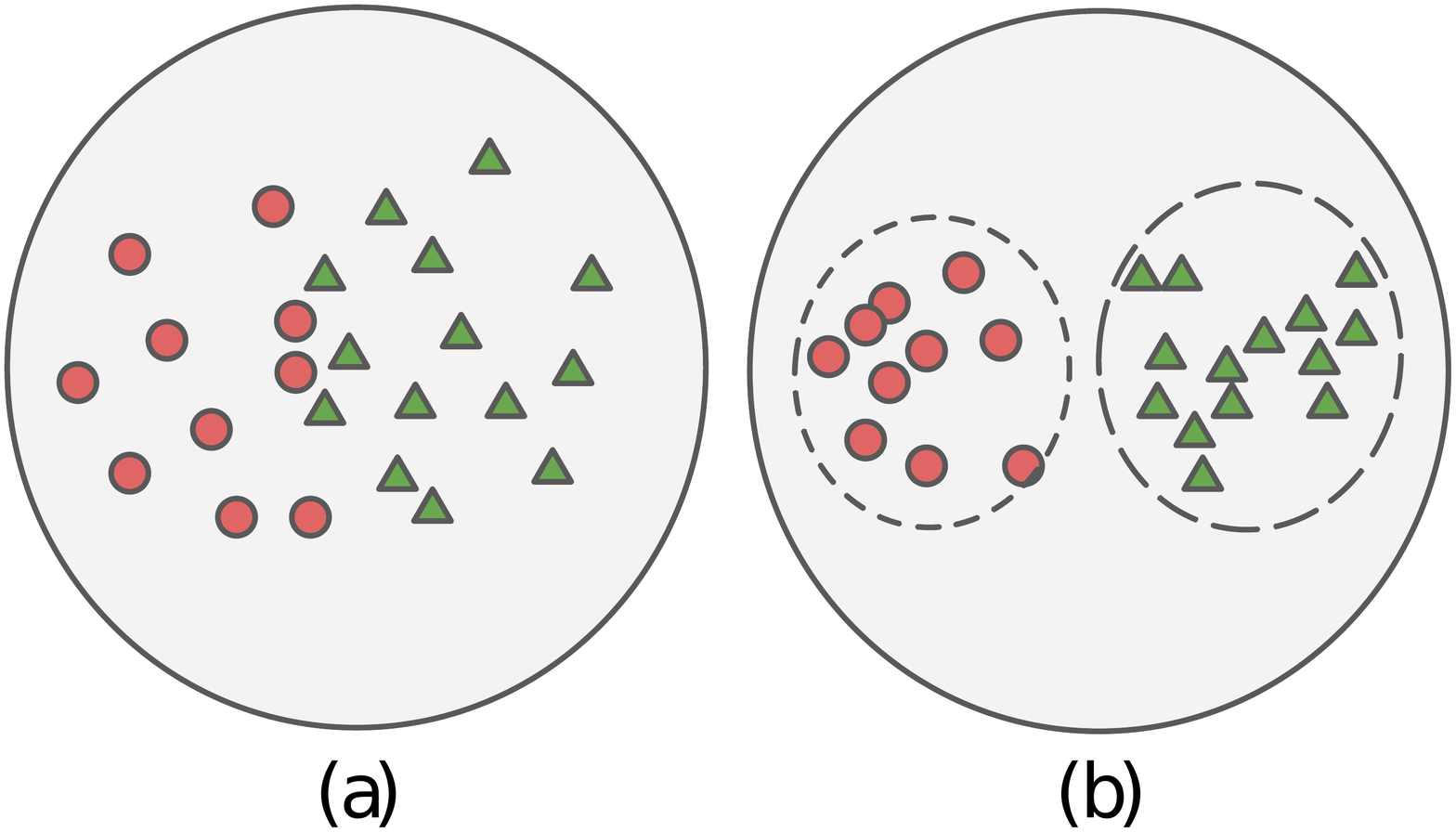}
    \vskip-2ex
    \caption{Simplified representation of the backbone's feature space. Standard training as shown in (a) does not encourage the features to be robust. This is fine if the training and test data is similar. In our case, we need more robust features which can be generated by using our proposed training procedure. Applying our proposed Panoramic Robust Feature (PRF) framework generates more robust features (b) which improve the performance on panoramic images.}
    \label{fig:robust}
\vskip-3ex
\end{figure}
\section{Related Work}

\subsection{Panoptic segmentation}

Semantic- and instance-specific segmentation have progressed greatly in the last years thanks to the architectural advances of deep models~\cite{long2015fully}\cite{he2016deep}\cite{he2017mask}.
The Fully Convolutional Network (FCN)~\cite{long2015fully} tackles semantic segmentation in an end-to-end way, followed by encoder-decoder architectures~\cite{badrinarayanan2017segnet}\cite{zhao2017pyramid}\cite{chen2017deeplab} that significantly improve segmentation performance by aggregating multi-scale contextual information.
Mask R-CNN~\cite{he2017mask} simultaneously addresses detection and segmentation using bounding box proposals, forming the basis of many current state-of-the-art instance segmentation networks.
The recently introduced panoptic segmentation task~\cite{kirillov2019panoptic} unifies semantic- and instance segmentation in a single perception system, allowing to recognize both things and stuff in urban driving scenes, which are of important relevance for autonomous vehicles.
A multitude of works~\cite{porzi2019seamless}\cite{de2019single}\cite{wang2020max} have presented panoptic segmentation by following a universal framework, showing the significance of the new task to the field.
Among these approaches, some develop based on state-of-the-art instance segmentation methods like Panoptic FPN~\cite{kirillov2019pfpn} extending Mask R-CNN with a semantic branch, whereas others enhance semantic segmentation architectures like
MaX-DeepLab~\cite{wang2020max} extending DeepLab~\cite{chen2017deeplab}.
Seamless-Scene-Segmentation~\cite{porzi2019seamless} directly incorporates an instance segmentation head following Mask R-CNN and a light-weight DeepLab-style semantic segmentation head.
Most recently, there are single-path architectures without using separate branches like DETR~\cite{carion2020end}, Panoptic FCN~\cite{li2020fully} and SPINet~\cite{hwang2020single}.
In this work, we build on seamless segmentation in an efficient setting, considering that fast responses are critical for autonomous driving systems.

\subsection{Panoramic segmentation}

Modern scene segmentation approaches are mostly designed to work with pinhole images on mainstream datasets like Cityscapes~\cite{cordts2016cityscapes} and Mapillary Vistas~\cite{neuhold2017mapillary}.
To enlarge the Field of View (FoV), early surrounding perception systems are based on fisheye images or multiple cameras~\cite{narioka2018understanding}\cite{deng2019restricted}. Motivated by the prospect of attaining wide-angle semantic perception with a single camera, recent works~\cite{zhang2019orientation}\cite{sekkat2020omniscape} build directly on this modality, relying on synthetic collections that are far less diverse than pinhole databases~\cite{cordts2016cityscapes}\cite{neuhold2017mapillary}.
In contrast, Panoramic Annular Semantic Segmentation (PASS)~\cite{yang2019pass} re-uses knowledge in pinhole data to produce robust models suitable for 360$^\circ$ images.
DS-PASS~\cite{yang2019ds} further augments with a detail-sensitive design by using attention connections.
In~\cite{yang2021context}, context-aware omni-supervised models are taken to the wild, fulfilling panoramic semantic segmentation in a single pass with enhanced generalizability.
However, existing wide-FoV systems only render semantic- or instance-specific segmentation.
To achieve a unified and comprehensive scene perception, this work first addresses panoramic panoptic segmentation, extending Seamless-Scene-Segmentation~\cite{porzi2019seamless} with a contrastive learning regimen that intertwines pixel-wise consistency propagation~\cite{xie2020propagate} for robust segmentation across pinhole- and panoramic imagery.

\subsection{Unsupervised visual representation learning}
Currently, the most appealing approaches for learning representations without labels are unsupervised contrastive learning~\cite{wu2018unsupervised}\cite{he2020momentum} tasks. Contrastive training methods learn visual representations in a discriminative way by contrasting similar, positive pairs against dissimilar, negative pairs, which is promising to yield generalized features for robust predictions in previously unseen domains. The training pairs are often generated from augmented views of image samples, and thereby the previous methods are mostly designed for object classification tasks, which does not ensure more accurate pixel-wise segmentation~\cite{he2019rethinking}. Different from previous works, we aim to develop a contrastive learning regimen for pixel-level tasks. Taking the wide FoV of omnidirectional data into consideration, in this paper, we propose a learning framework for fine-grained panoptic segmentation operating on panoramic images. Although a few latest contrastive training methods~\cite{wang2020dense}\cite{wang2021exploring}, concurrent to ours, also address dense prediction tasks, we step further to open 360$^\circ$ scenes and explore the generalization effects.

\section{Proposed Framework}
\label{Sec: Proposed Framework}
\begin{figure*}
    \centering
    \includegraphics[trim=150 175 125 145, clip, width=\textwidth]{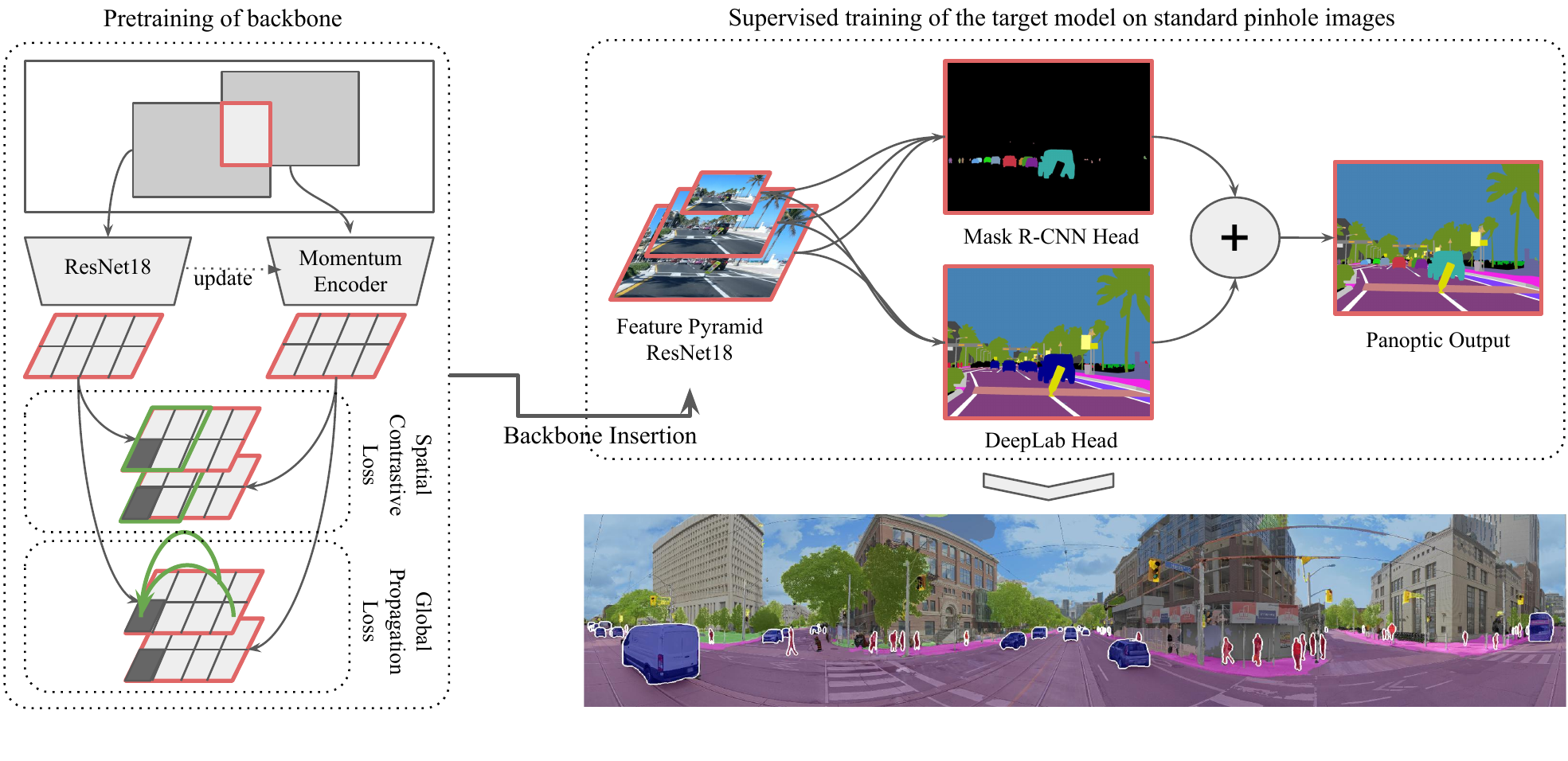}
    \vskip-4ex
    \caption{Overview of the proposed Panoramic Robust Feature (PRF) framework consisting of a pretraining and a standard supervised training step. During the pretaining, we encourage the backbone of the model to learn more robust features. The adapted backbone is then inserted into the target model which can be trained on standard pinhole images. Due to the more robust features, we boost the performance for the desired panoramic panoptic image segmentation. During pretraining, we optimize for spatial sensitivity and global propagation. Spatially close pixels (green frame) are pushed to have similar representations and globally similar pixels are propagated onto each other (green arrows) and encouraged to be consistent across the two frames.}
    \label{fig: framework overview}
    \vskip-3ex
\end{figure*}

The proposed Panoramic Robust Feature (PRF) framework consists of two steps: A short pretraining step which is responsible to initialize the network's backbone with a robust feature representation. The difference between the feature space of a standard backbone and our pretrained backbone is visualized in Fig.~\ref{fig:robust}. The pretraining is followed by standard supervised training on available labeled datasets. We show that even though the pretraining step only adds minimal overhead compared to the main supervised training, it is able to boost the model performance on data from a different domain. This is the desired effect for our application since we will leverage large already available datasets such as Cityscapes~\cite{cordts2016cityscapes} and will perform prediction on the previously unseen panoramic data domain.

The proposed training procedure is shown in Fig.~\ref{fig: framework overview}. Driven by the need for data independent model performance, we aim for an increase of the robustness of the features. Our proposed strategy involves the use of contrastive learning techniques. Specifically, we consider that panoptic segmentation is a pixel-level task, and thereby propose to leverage pixel-level contrastive learning~\cite{xie2020propagate} and adapt it for our 360$^\circ$ panoramic segmentation, which unfolds as very beneficial according to our experiments in Section~\ref{Sec: Experiments}.

After pretrainng, the supervised training procedure of the
target model can be executed on labeled pinhole images.
For our experiments, we use an efficient modification of the state-of-the-art Seamless-Scene-Segmentation model~\cite{porzi2019seamless} as our target model for panoptic image segmentation due to its simplicity and versatility. Following its general architecture, we replace the standard ResNet50~\cite{he2016deep} backbone with a ResNet18~\cite{he2016deep} network. This light-weight version is more efficient during both training and inference and is thus more suitable for mobile intelligent vehicles with only limited computational resources. 
In the following, we describe our learning processes in detail.

\subsection{The unsupervised pretraining phase}
\label{Ss: target model training}
We start by cropping two frames from the same image and apply standard image augmentation techniques to each of those frames. Those include random flipping, color jittering, random gray scale conversion, random solarization and normalization. We feed each of the augmented views to one of two pipelines: One frame is passed through the encoder (e.g. ResNet18) with a projection head on top while the other one is passed through a momentum encoder~\cite{he2020momentum}. After each training step $t$, the weights $\theta_t$ of the momentum encoder are updated according to Equ.~\ref{Eq: Moment Encoder weight update}, where $\beta$ is a multiplicative factor and $\theta^{enc}_t$ denotes the weights of the regular encoder at time $t$. 
\begin{equation}
    \theta_t = \theta_{t-1}\times\beta + (1-\beta)\times\theta^{enc}_t
    \label{Eq: Moment Encoder weight update}
\end{equation}
The momentum encoder can thus be seen as an exponentially weighted average of the regular encoder. Both networks should thus produce similar output features for the same input, which is the main idea upon which the training losses rely. The projection head is a small network consisting of a $1\times 1$ convolutional layer followed by a batch normalization and a ReLU layer before a final $1\times 1$ convolutional layer projects the features into the target dimension. We use $2048$ as the latent space dimensionality and $256$ output channels following~\cite{xie2020propagate}.
Upon the two views and the two different representations, two different types of consistencies are demanded with very different effects on the learned representations. We briefly revisit those and refer the interested reader to the original paper~\cite{xie2020propagate}.

Spatial contrastive loss: Considering the wide-FoV of panoramic data, this loss is designed to cluster the representations of spatially close pixels (indicated by the green frame in Fig.~\ref{fig: framework overview}). The basic idea behind this loss is that two feature vectors $x_i$ and $x'_i$ computed by the encoder and momentum encoder respectively which correspond to the same pixel $p_i$ in the original image should be similar. 
This idea is extended to enforcing similarities across the feature vectors $x$ and $x'$ for not just the same but spatially close pixels.
More precisely, if we consider the first frame $F$ and pixel $p_i$ located within that first frame, $\tilde{F}^P_i$ are the pixels in the second frame which are spatially close to pixel $p_i$ measured in the original image space, whereas $\tilde{F}^N_i$ are those pixels further apart. $\tau$ is a normalization parameter which we set to $0.3$ as suggested in \cite{xie2020propagate}. The loss in Equ.~\ref{Eq: Spatial contrastive loss} is first averaged across all pixels of the same frame and finally the losses for the first and the second frame are averaged.
\begin{equation}
    L_{s}(p_i) = -\log\frac{\sum_{j \in \tilde{F}^P_i}e^{\frac{cos(x_i,x'_j)}{\tau}}}{\sum_{j \in \tilde{F}^P_i}e^{\frac{cos(x_i,x'_j)}{\tau}} + \sum_{k \in \tilde{F}^N_i}e^{\frac{cos(x_i,x'_k)}{\tau}}}
    \label{Eq: Spatial contrastive loss}
\end{equation}

Global propagation loss:
Whereas the spatial contrastive loss forces spatially close pixels to have similar representations which results in spatial sensitive representations, a second desirable goal is to prefer smooth outputs of the network over very fragmented labels. This effect can be achieved by projecting the features of globally similar pixels onto the current pixel which introduces a smoothing effect, capturing long-range dependencies that stretch across the whole FoV. Equ.~\ref{Eq: Pixel Projection} shows the formula to calculate a smoothed feature $x_i^{smooth}$ of $x_i$ as a weighted sum of the projections $g(x)$ of any feature $x_j$ generated from all pixels within the same frame $F$ in which the weights are determined by the similarity between $x_i$ and $x_j$. $g(x)$ is computed via a $1\times 1$ convolutional layer which keeps the number of channels constant. 
\begin{equation}
    x_i^{smooth}=\sum_{j \in F}\max(\cos(x_i,x_j),0)^2*g(x_j)
    \label{Eq: Pixel Projection}
\end{equation}
Finally, we encourage consistency between the features generated by spatially close pixels $i$ and $j$ in the original image space representing the same information that was computed by the two pipelines.
\begin{equation}
L_{GloPro} = -\cos(x_i^{smooth},x'_j)-\cos(x_j^{smooth},x'_i)
\label{Eq: Pixel Pro Loss}
\end{equation}

We find that the additive combination of those two losses is very beneficial for our desired application. Compared to the baseline model which is not pretrained but only trained using the supervised approach described in Section~\ref{Ss: Supervised pretraining}, we observe the expected smoothness of the labels which remarkably improves the performance of the classes that substantially differ in location and looks between the training and the test set. Examples for this very common behaviour can be seen in Fig.~\ref{fig: qualitative results}.

\subsection{The supervised training phase}
\label{Ss: Supervised pretraining}
After the previously described pretraining, we insert the pretrained backbone into the target model which is then trained using the model specific training procedure. In our setting, we choose the Seamless-Scene-Segmentation model~\cite{porzi2019seamless} which applies a feature pyramid~\cite{lin2017feature} on the ResNet18 backbone. The model then feeds multiple scales from the pyramid to the instance segmentation and the semantic segmentation head. The semantic segmentation head consisting of a DeepLab~\cite{chen2017deeplab} inspired light-weight segmentation network is optimized using a binary cross entropy loss. The instance segmentation head consists of a Mask R-CNN Model which is trained using several bounding box detection losses and the mask segmentation loss~\cite{he2017mask}\cite{porzi2019seamless}. 

The panoptic output is achieved by a final fusion step~\cite{kirillov2019panoptic}\cite{porzi2019seamless} which merges the output of the instance segmentation head with the semantic segmentation head. The quality of the output is calculated according to the panoptic quality measure~\cite{kirillov2019panoptic}.

Finally, we obtain panoramic panoptic segmentation in a single shot with the trained efficient model that can be deployed in intelligent vehicle systems, delivering a complete and robust surrounding understanding. 

\section{Experiments}
\label{Sec: Experiments}
\subsection{Datasets}
Following WildPASS~\cite{yang2021context}, we collect 40 panoramic images from 40 cities around the world, in order to build a diverse evaluation set for the desired panoramic panoptic segmentation. Each of the panoramic images (at $400\times2048$, $70^\circ{\times}360^\circ$) is annotated with the most relevant classes for street scene understanding, which are the stuff classes \textit{street} and \textit{sidewalk} as well as the thing classes \textit{person} and \textit{car}. The proposed Wild Panoramic Panoptic Segmentation (WildPPS) dataset is used to assess the model performance on the panoramic images. For the supervised training, we rely on large annotated already existing pinhole image datasets such as Cityscapes~\cite{cordts2016cityscapes} and Mapillary Vistas~\cite{neuhold2017mapillary}. Cityscapes is a street scenery dataset consisting of 5000 fine annotated images spit into 2975/500/1525 images forming the train, validation and test set respectively. The images are captured from 50 cities in Germany and taken under comparable weather and lighting conditions. The Cityscapes Coarse dataset adds 19998 images with coarse annotations. 
Vistas is a more diverse and larger street scene dataset consisting of 25k fine annotated images split into 18k/2k/5k train, validation and test sets captured from all over the world.

\subsection{Experimental Setup}

Due to its wide acceptance within the community, we choose the Cityscapes dataset as the primary dataset for our experiments. 
All of the following experiments were designed with an intelligent vehicle application in mind. We thus restricted our hardware usage to a maximum of two GPUs for training and testing, in order to work with a more realistic setting that could be deployed on a mobile system. 

As a baseline model, we use a non-adapted model for which we skip the pretaining phase and directly perform a supervised training routine on the Cityscapes training dataset. We train for 52 epochs using the training setting described in the Seamless-Scene-Segmentation paper~\cite{porzi2019seamless} with two modifications: we resize the images to $512\times1024$ pixels to meet our hardware requirements and reduce the initial learning rate of $1e\text{-}2$ by a factor of $10$ after 25k and 35k iterations due to observed loss saturation. In order to receive a fair comparison, we apply the exact same supervised training procedure to every backbone modification described in the following.

In order to assess the performance of the framework, we experiment with multiple optimizers and learning rates which is described in the following section. The pretraining is performed using a single GPU on the 2975 training images of Cityscapes unless stated otherwise. After the pretraining, we insert the ResNet18 into the target model and perform the aforementioned supervised training routine.

\subsection{Experimental Results}
\subsubsection{Experiments on WildPPS}
The best performing pretrain procedures for WildPPS are obtained by using a relatively large batch size of $100$ and a large base learning rate of $0.4$ that is reduced according to a cosine annealing schedule with restarts every $30$ epochs for a total of $90$ epochs. Gradient updates are performed using the LARS~\cite{you2017large} optimizer due to its suitability for large batch sizes. This setting does outperform a more standard setting in which we used SGD with momentum and a moderate learning rate of $1e\text{-}3$. Using this procedure we manage to improve more than $4\%$ over the baseline model.

\textbf{Pretrain loss modifications:} We find the Global Propagation Loss to be slightly more beneficial for our application since the best overall results for pretraining on Cityscapes train dataset are obtained when optimizing a skewed pretraining loss $L_{pretrain}=L_s+\alpha\times L_{GloPro}$ with $\alpha=2$, which puts more weight on the Global Propagation Loss. Experiments using the aforementioned loss modification are marked as LARS$^*$, while all other experiments are performed using equal weights for the two losses. The re-weighted loss allows us to boost the model performance even above the best improvements using the standard loss on the WildPPS dataset. Using the modified loss we are able to improve the baseline model well above $5\%$.

\textbf{Pretrain data modification:}
An additional benefit can be obtained by using more data for the pretraining phase. 
Cityscapes Coarse offers almost 20k images under the same distribution. 
To keep training times comparable, we train on this larger image database using the same learning rate schedule than before, but we skip the cosine annealing restarts and finish training after 30 epochs. We point out that even though we perform unsupervised pretraining on Cityscapes Coarse, we still do the supervised training on the standard Cityscapes training set. Using this setting, we obtain the best performing model which improves an impressing $5.1\%$ over the baseline model and $12.4\%$ in the stuff class. The qualitative improvements of this model can be seen in Fig.~\ref{fig: qualitative results}.

We refer the reader to Table~\ref{table_panoptic} for more details on the experiments on WildPPS. For more details on the pretraining settings, we refer to our used guidelines for pixel-level contrastive learning~\cite{xie2020propagate}.

\begin{table}[t]
\setlength{\tabcolsep}{2.0pt}
    \centering
    \caption{Comparison of results using the proposed framework on WildPPS evaluation set and Mapillary Vistas validation set.}
    \vskip-1ex
    \begin{tabular}{|c|c|c|c|}
    \hline
        \textbf{Pretrain Setting} &  \textbf{PQ Stuff} & \textbf{PQ Things} & \textbf{PQ}\\
        \hline
        \hline
        \multicolumn{4}{|c|}{WildPPS evaluation set}\\
        \hline
         Baseline Model (no pretraining) & 57.9\% & 54.2\% & 56.0\%\\
         \hline
         Pretrain Cityscapes SGD & 63.4\% & \textbf{55.8\%} & 59.6\% \\
         \hline
         Pretrain Cityscapes LARS$^{*}$ $(\alpha{=}0.5)$ & 64.1\% & 55.4\% & 59.8\%\\
         \hline
         Pretrain Cityscapes LARS $(\alpha{=}1)$ & 67.0\% & 53.2\% & 60.1\% \\
         \hline
         Pretrain Cityscapes LARS$^{*}$ $(\alpha{=}2)$ & 67.0\% & 55.2\% & 61.1\%\\
         \hline
         Pretrain Cityscapes LARS$^{*}$ $(\alpha{=}4)$ & 65.3\% & 53.1\% & 59.2\%\\
         \hline
         \hline
         \textbf{Pretrain Cityscapes Coarse LARS} & \textbf{70.3\%} & 52.6\% & \textbf{61.4\%} \\
         \hline
         \hline
         \multicolumn{4}{|c|}{Mapillary Vistas validation set}\\
         \hline
         \hline
         Baseline Model (no pretraining) & 38.6\% & 20.7\% & 30.2\%\\
         \hline
         \textbf{Pretrain Cityscapes SGD} & \textbf{41.7\%} & 20.7\% & \textbf{31.9\%}\\
         \hline
         Pretrain Cityscapes LARS & 40.2\% & 21.2\% & 31.3\% \\
         \hline
         Pretrain Cityscapes LARS$^{*}$ $(\alpha{=}2)$ & 39.2\% & \textbf{21.8\%} & 31.1\%\\
         \hline
         Pretrain Cityscapes Coarse LARS & 38.4\% & 21.3\% & 30.4\%\\
         \hline
    \end{tabular}
    \label{table_panoptic}
    \vskip-3ex
\end{table}

\subsubsection{Experiments on Mapillary Vistas}
As our framework generates robust feature representations which are beneficial when train and test set differ, we designed an experiment in which a model is first trained on the Cityscapes training set before the model performance is measured on the Mapillary Vistas validation set.
In order to measure the Cityscapes trained model on the much richer Mapillary Vistas validation set, we map each of the Cityscapes classes to the corresponding Mapillary vistas class and ignore all classes for which the model does not produce predictions. The mapping between Cityscapes and Mapillary Vistas classes is straightforward with two exceptions: we map the Cityscapes classes \textit{rider} and \textit{train} to the Vistas classes \textit{motorcyclist} and \textit{other vehicle} respectively.
The findings are shown in Table~\ref{table_panoptic} and are in line with our expectations. Even though we find the order of best performing models to be different, each of the models outperforms the base model by a significant margin.  
Since the Cityscapes and the Vistas datasets are more alike than Cityscapes and WildPPS, we only see a smaller improvement over the baseline, which is however to be expected. 

\begin{table*}[t]
\setlength{\abovecaptionskip}{0pt}
\setlength{\belowcaptionskip}{0pt}
\caption{Per-class accuracy analysis in Intersection over Union (IoU) and mean IoU (mIoU)\protect\\on the public Panoramic Annular Semantic Segmentation (PASS) dataset~\cite{yang2019pass}.}
\vskip-1ex
\label{table_pass}
\begin{center}
\begin{tabular}{|c|c|c|c|c|c|c|c|}
\hline
{\textbf{Network}}&{\textbf{Car}}&{\textbf{Road}}&{\textbf{Sidewalk}}&{\textbf{Crosswalk}}&{\textbf{Curb}}&{\textbf{Person}}&{\textbf{mIoU}}\\
\hline
\hline
{SegNet~\cite{badrinarayanan2017segnet}}&{57.5\%}&{52.6\%}&{17.9\%}&{11.3\%}&{11.6\%}&{3.5\%}&{25.7\%}\\
\hline
{PSPNet (ResNet50)~\cite{zhao2017pyramid}}&{76.2\%}&{67.9\%}&{34.7\%}&{19.7\%}&{27.3\%}&{22.6\%}&{41.4\%}\\
\hline
{DenseASPP (DenseNet121)~\cite{yang2018denseaspp}}&{65.8\%}&{62.9\%}&{30.5\%}&{8.7\%}&{23.0\%}&{8.7\%}&{33.3\%}\\
\hline
{DANet (ResNet50)~\cite{fu2019dual}}&{70.0\%}&{67.8\%}&{35.9\%}&{21.3\%}&{12.6\%}&{25.9\%}&{38.9\%}\\
\hline
\hline
{ENet~\cite{paszke2016enet}}&{59.4\%}&{59.6\%}&{27.1\%}&{16.3\%}&{15.4\%}&{8.2\%}&{31.0\%}\\
\hline
{CGNet~\cite{wu2021cgnet}}&{65.2\%}&{56.9\%}&{23.7\%}&{3.8\%}&{11.2\%}&{21.4\%}&{30.4\%}\\
\hline
{ERFNet~\cite{romera2018erfnet}}&{70.0\%}&{57.3\%}&{25.4\%}&{22.9\%}&{15.8\%}&{15.3\%}&{34.3\%}\\
\hline
{PSPNet (ResNet18)~\cite{zhao2017pyramid}}&{64.1\%}&{67.7\%}&{31.2\%}&{15.1\%}&{17.5\%}&{12.8\%}&{34.8\%}\\
\hline
{ERF-PSPNet~\cite{yang2019pass}}&{71.8\%}&{65.7\%}&{32.9\%}&{29.2\%}&{19.7\%}&{15.8\%}&{39.2\%}\\
\hline
{ERF-PSPNet (Omni-Supervised)~\cite{yang2021context}}&{81.4\%}&{\textbf{71.9\%}}&{\textbf{39.1\%}}&{24.6\%}&{26.4\%}&{44.1\%}&{47.9\%}\\
\hline
{SwiftNet~\cite{orvsic2019defense}}&{67.5\%}&{70.0\%}&{30.0\%}&{21.4\%}&{21.9\%}&{13.7\%}&{37.4\%}\\
\hline
{SwaftNet~\cite{yang2019ds}}&{76.4\%}&{64.1\%}&{33.8\%}&{9.6\%}&{26.9\%}&{18.5\%}&{38.2\%}\\
\hline
\hline
{\textbf{Ours}}&{\textbf{84.7\%}}&{68.8\%}&{37.5\%}&{\textbf{50.2\%}}&{\textbf{27.4\%}}&{\textbf{61.4\%}}&{\textbf{55.0\%}}\\
\hline
\end{tabular}
\end{center}
\vskip-3ex
\end{table*}

\subsection{Comparison to previous models on PASS}
\begin{figure*}
    \centering
    \includegraphics[trim=75 170 85 155, clip, width=0.99\textwidth]{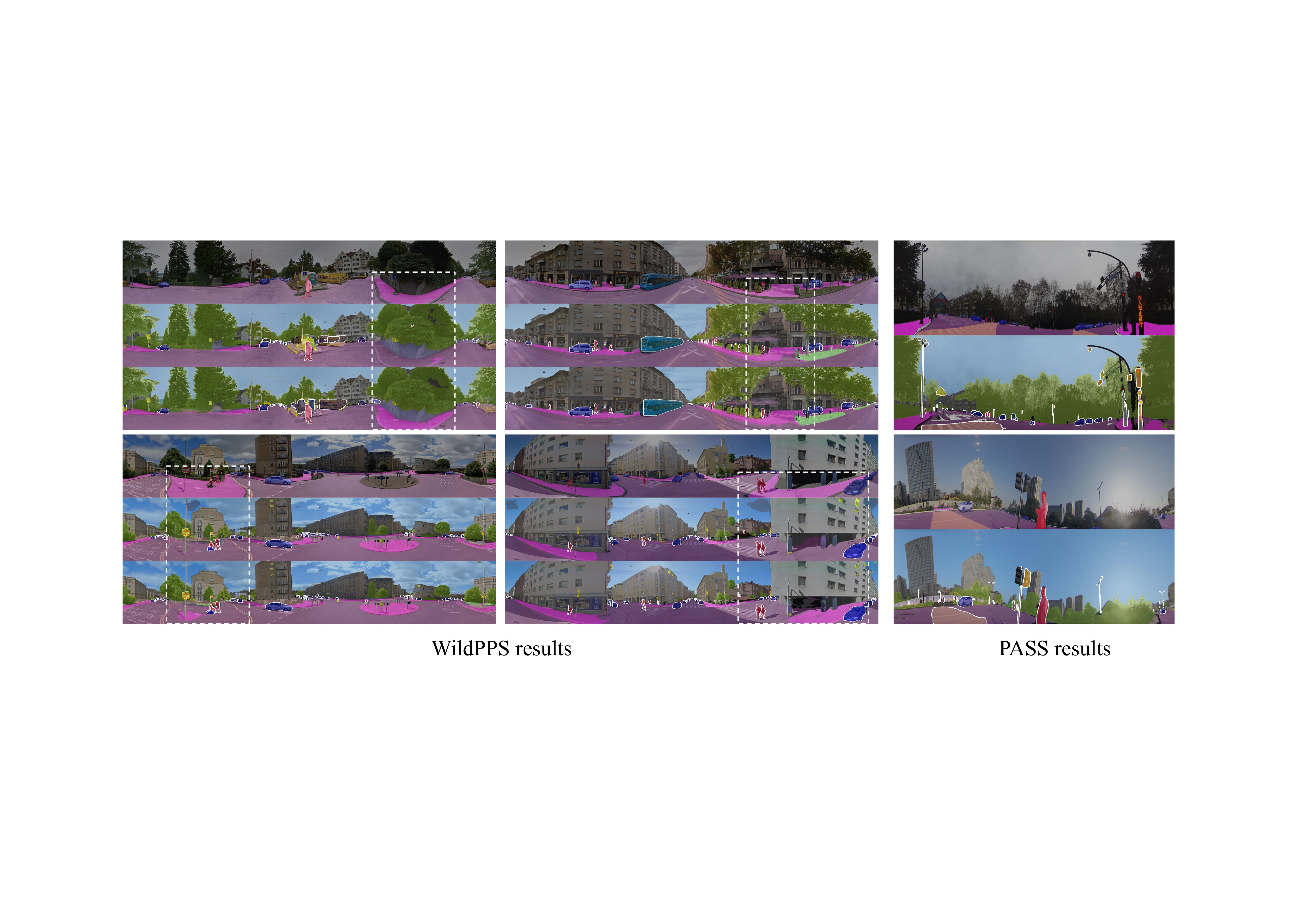}
    \vskip-2ex
    \caption{Qualitative results on our WildPPS dataset and on the public PASS dataset~\cite{yang2019pass}. For WildPPS, the first row of the image triplets shows the original image overlaid by the ground truth. The second row is the baseline model, whereas the the third row is the adapted model according to our proposed framework. We highlight stunning improvements within the white boxes. It can be seen that the Panoramic Robust Feature framework removes positional priors obtained by standard supervised training. In particular, the sidewalk is affected by this behaviour. Our robust features mitigate those effects allowing a position-independent segmentation required for high-quality panoramic panoptic predictions. 
    For the PASS dataset used for model comparison, we show image tuples of original images overlaid by the ground truth and our model predictions. Best viewed on a screen and with zoom.}
    \label{fig: qualitative results}
    \vskip-3ex
\end{figure*}

As we are the first to achieve panoptic segmentation on panoramic images, we compare our model to previous works which achieved semantic segmentation on the publicly available Panoramic Annular Semantic Segmentation (PASS) dataset~\cite{yang2019pass}. 
These networks, experimented by~\cite{yang2021context}, include state-of-the-art models SegNet~\cite{badrinarayanan2017segnet}, PSPNet50~\cite{zhao2017pyramid}, DenseASPP~\cite{yang2018denseaspp} and DANet~\cite{fu2019dual}, as well as efficiency-oriented models ENet~\cite{paszke2016enet}, CGNet~\cite{wu2021cgnet}, ERFNet~\cite{romera2018erfnet}, PSPNet18~\cite{zhao2017pyramid}, ERF-PSPNet~\cite{yang2019pass}, SwiftNet~\cite{orvsic2019defense} and SwaftNet~\cite{yang2019ds}. They all view the wide-FoV panoramic image as a single input without separating it into multiple segments. For a fair comparison, we measure the Intersection over Union (IoU) of the different classes and compare them to our robust Seamless-Scene-Segmentation model. Besides restricting ourselves to only comparing the semantic output of our model, we want to point out that our efficient modification of the Seamless-Scene-Segmentation model is also among the most efficient models within the comparison.
We apply our proposed framework to train a ResNet18 backbone followed by supervised training on Mapillary Viastas. We follow the supervised training procedure described in the previous section for a total of $23$ epochs. Starting with a base learning rate of $1e\text{-}2$, we reduce by a factor of $10$ after 95k and 125k iterations due to observed loss saturation. The results are shown in Table~\ref{table_pass}. Compared to the previous state-of-the-art networks, we surpass them by a large margin by using our proposed framework. In particular, the safety-critical class person is well addressed by our model, surpassing the second best model by $17.3\%$ in IoU. Qualitative results can be found in Fig.~\ref{fig: qualitative results}. 

\section{Conclusion}
In this work, we have introduced the novel task of panoramic panoptic image segmentation. Coming from semantic segmentation on classic pinhole images, panoramic panoptic segmentation extends the view from a narrow field of view to a 360$^\circ$ view and from a simpler scene understanding to a more complete scene understanding that differentiates between different instances of countable objects. We point out the importance of both additional sources of information and acknowledge the lack of available annotated panoramic images to train complex models on them. With the proposed Panoramic Robust Feature framework based on unsupervised contrastive learning, our unified, seamless 360$^\circ$ segmentation achieves improved robustness in open, previously unseen panoramic domains, significantly elevating state-of-the-art single-pass holistic scene understanding.

We hope that our work sparks the interest of the community in panoramic panoptic image segmentation. In the future, we plan to explore the beneficial effects of our framework on more target models. We are especially interested in constructing single-branch panoptic models and how an end-to-end unsupervised pretraining on them compares to inserting a pretrained backbone. Another possible research direction is to include a self-attention layer into our model, which is very beneficial for our task of panoramic panoptic segmentation as it allows a direct information exchange among all pixels.

\bibliographystyle{IEEEtran}
\bibliography{bib.bib}

\end{document}